\documentclass[psfig,twocolumn,10pt]{asme2e_JMR}
\usepackage{amsfonts}
\usepackage{graphicx}
\usepackage{epstopdf}
\usepackage{amsmath, amssymb, bm}
\usepackage[fixamsmath,disallowspaces]{mathtools}
\usepackage{graphicx}
\usepackage{color}
\usepackage{subfig}
\usepackage{amssymb}
\usepackage{amsmath}
\usepackage{url}
\usepackage{array}
\usepackage{algorithm}
\usepackage{algpseudocode}
\usepackage{enumerate}
\usepackage{todonotes}
\usepackage{wallpaper}
\usepackage{epsfig}
\usepackage{caption}
\usepackage{epstopdf}
\usepackage{fixmath}

\newtheorem{proposition}{Proposition}

\begin{document}

\parindent0pt \setcounter{topnumber}{9} \setcounter{bottomnumber}{9} %
\renewcommand{\textfraction}{0.00001}

\renewcommand {\floatpagefraction}{0.999} \renewcommand{\textfraction}{0.01} %
\renewcommand{\topfraction}{0.999} \renewcommand{\bottomfraction}{0.99} %
\renewcommand{\floatpagefraction}{0.99} \setcounter{totalnumber}{9}

\title{ 
Identification of Non-transversal Motion Bifurcations of Linkages
\vspace{-8mm}
} 
\author{
	{Andreas M\"uller}
	\affiliation{Johannes Kepler University, Linz, Austria\\
	a.mueller@jku.at}
\and
	{P.C. L\'opez-Custodio}
	\affiliation{King's College London, UK\\
	pablo.lopez-custodio@kcl.ac.uk}
\and
	{J.S. Dai}
	\affiliation{King's College London, UK\\
	jian.dai@kcl.ac.uk}
} 

\maketitle 

\begin{abstract}
\textit{Abstract--} The local analysis is an established approach to the
study of singularities and mobility of linkages. Key result of such analyses
is a local picture of the finite motion through a configuration. This
reveals the finite mobility at that point and the tangents to smooth motion
curves. It does, however, not immediately allow to distinguish between
motion branches that do not intersect transversally (which is a rather
uncommon situation that has only recently been discussed in the literature).
The mathematical framework for such a local analysis is the kinematic
tangent cone. It is shown in this paper that the constructive definition of
the kinematic tangent cone already involves all information necessary to
distinguish different motion branches. A computational method is derived by
amending the algorithmic framework reported in previous publications.


\end{abstract} 

\section{Introduction}

A motion of a mechanism corresponds to a curve in its configuration space
(c-space) $V\subset {\mathbb{V}}^{n}$. In a generic configuration, i.e. at a
generic point $\mathbf{q}\in V$, the c-space is a smooth manifold, and its
tangent space provides a proper first-order description of possible motions,
i.e. of finite curves through $\mathbf{q}$. This fails in c-space
singularities. The analysis of singularities of mechanisms has traditionally
been focused on those that are due to bifurcations of the c-space. At a
bifurcation singularity several 'motion branches' intersect. The latter can
be of different dimensions, as in case of kinematotropic mechanisms \cite%
{Wohlhart1996}.

The analysis of c-space singularities can be approached from a global
perspective with the aim to obtain an exhaustive description of all
singularities. To this end, an algebraic description of the loop constraints
is used, e.g. in terms of (dual) quaternions or Study parameters, and the so
defined algebraic variety is analyzed with algorithms from computational
algebraic geometry. Such investigations were reported by several authors,
e.g. \cite{Kong-MMT2018}. The basic problem of this method is the
non-polynomial complexity of the involved algorithms. The computational
effort further depends on the particular algebraic formulation of the
constraint equations. This issue is mitigated by the linear implicitization
algorithm \cite{WalterHusty2010LIA,HustyWalter2019,Arponen2008}. Yet, a
global analysis remains possible for specific examples only. A local
analysis, on the other hand, aims at identification of the local geometry of
the c-space. Various such approaches were reported \cite%
{Bustos2012,Chen2011,Kieffer94,Lerbet1999,Rico1999,JMR2018}. Their common
basis is a local approximation of finite motions through a given (singular)
configuration. A systematic higher-order local analysis method for general
linkages was proposed in \cite{JMR2018}. The key object is the kinematic
tangent cone \cite{CISMMueller2019,Lerbet1999}. The closed form and
recursive relations for the higher-order loop constraints reported in \cite%
{CISMMueller2019,Mueller-MMT2019} ensure its practical applicability to
general multi-loop linkages. It must be noticed that all local methods
attempt an approximation of the finite motions through a given
configuration, rather than of the c-space geometry (see section \ref%
{secKinTangCone}).

The local analysis determines tangents to smooth curves in $V$, i.e. the
possible 'directions' of finite motions. It thus allows to identify the
finite local mobility (or mobilities in case of kinematotropic linkages) at
bifurcation singularities, which is an important result for studying the
mobility and singularities of linkages. In addition to the finite
mobility/mobilities, one is usually interested in distinguishing the motion
branches intersecting at a bifurcation singularity. The proposed local
analysis methods fail to distinguish motion branches if they do not
intersect transversally, however, in which case different motion branches
possess the same tangents. This problem is addressed in this paper. Clearly,
the tangential aspects are not sufficient, but higher-order information
(curvature etc.) must be used. It will be shown that all necessary
informations are already available from the constructive definition of the
kinematic tangent cone.

There is a plethora of publications on transversal bifurcations, e.g. \cite%
{HerveLi2009,Gogu2011,ParkKim1999,ZlatanovBonevGosselin2002,Feng2017}.
However, to the authors' knowledge, non-transversal bifurcations were not
reported in the literature except in \cite{PabloMMT2020}, where a systematic
construction of 1-DOF single-loop linkages with tangentially intersecting
motion branches was presented.

In this paper, a mathematical framework along with a computational method is
proposed for the higher-order local analysis of non-transversal
singularities of linkages, i.e. mechanical systems that can be modeled as an
assembly of rigid links and lower-pair joints. It builds upon the general
method for higher-order singularity analysis reported in \cite%
{JMR2018,CISMMueller2019}. The computational steps exploit the algebraic
formulations of higher-order kinematics that were summarized in \cite%
{Mueller-MMT2019}, and for which a Mathematica$^{^{\copyright }}$
implementation is available at \cite{Mueller-Mendeley2019}. The paper is
organized as follows. In section \ref{secConstraints} the formulation of
loop constraints in terms of joint screw coordinates for single-loop as well
as for multi-loop linkages is recalled. The local analysis method and the
concept of kinematic tangent cone is summarized in section \ref{secLocAnal}.
With these preliminaries, the approach to identify non-transversal
bifurcations is introduced in section \ref{secLocAnalTangent}.

\section{Kinematic Constraints for Multi-Loop Linkages%
\label{secConstraints}%
}

Consider first a single-loop linkage comprising $n$ helical joints
(revolute, prismatic, screw joints). The geometric constraints of the
kinematic loop are expressed as $f\left( \mathbf{q}\right) =\mathbf{I}$
where the constraint mapping $f:{\mathbb{V}}^{n}\rightarrow SE\left(
3\right) $ is defined by the product of exponentials%
\begin{equation}
f\left( \mathbf{q}\right) =\exp \left( q_{1}\mathbf{Y}_{1}\right) \exp
\left( q_{2}\mathbf{Y}_{2}\right) \cdot \ldots \cdot \exp \left( q_{n}%
\mathbf{Y}_{n}\right) .  \label{f}
\end{equation}%
Therein, $\mathbf{Y}_{i}\in {\mathbb{R}}^{6}\cong se\left( 3\right) $ is the
screw coordinate vector of joint $i$ represented in a global reference
frame, and $q_{i}$ is the joint variable (angle, translation). The c-space
of the single-loop linkage is then the analytic variety $V:=\left\{ \mathbf{q%
}\in {\mathbb{V}}^{n}|f\left( \mathbf{q}\right) =\mathbf{I}\right\} $.

For a multi-loop linkage, topologically independent fundamental cycles (FC),
denoted $\Lambda _{1},\ldots ,\Lambda _{\gamma }$, can be identified for
which loop closure constraints are defined \cite{DavisMMT2015,Robotica2018}.
The kinematic topology is represented by a topological graph $\Gamma $. A
directed graph $\vec{\Gamma}$ is introduced where directions of the edges
are according to how relative motions are measured, which is also called the
'polarity' of the joints \cite{Featherstone2008}. In each FC there is
exactly one edge from the co-tree $\mathcal{H}$ (the complement of the
spanning tree $\mathcal{G}$). An ordering of edges within a FC is defined
according to the order in which edges are visited when traversing the FC
starting from the co-tree edge. This relation is denoted with ${<_{l}}$.

The geometric loop constraints for the FC $\Lambda _{l}$ is $f_{l}\left( 
\mathbf{q}\right) =\mathbf{I}$, where%
\begin{align}
f_{l}\left( \mathbf{q}\right) :=& \exp (\sigma _{l,\underline{l}}q_{%
\underline{l}}\mathbf{Y}_{\underline{l}})\exp (\sigma _{l,i}q_{i}\mathbf{Y}%
_{i})\exp (\sigma _{l,j}q_{j}\mathbf{Y}_{j})\cdot \ldots  \notag \\
& \ldots \cdot \exp (\sigma _{l,{\overline{l}}}q_{\overline{l}}\mathbf{Y}_{{%
\overline{l}}}),\ {\underline{l}}\,{<_{l}}\,i\,{<_{l}}\,j\,{<_{l}}\,\ldots {%
<_{l}}\,\overline{l}\in \Lambda _{l}
\end{align}%
is the constraint mapping for $\Lambda _{l}$ \cite{CISMMueller2019}, in
which $\sigma _{l,i}\in \{-1,1\}$ indicates the 'orientation' of joint $i$
within the FC $\Lambda _{l}$. 
\begin{figure}[tb]
\begin{center}
a)~~~~~\hspace{-2ex}%
\includegraphics[width=9.0cm]{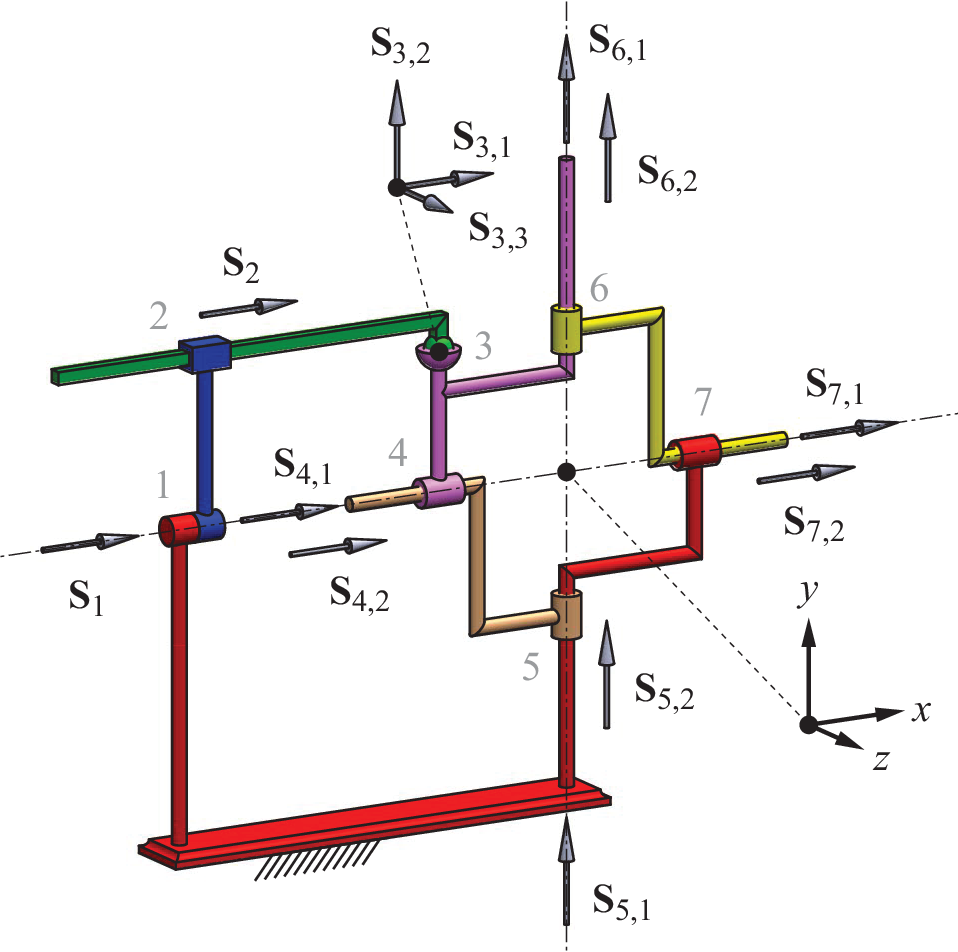}\\[%
0pt]
b)~~~\includegraphics[width=7cm]{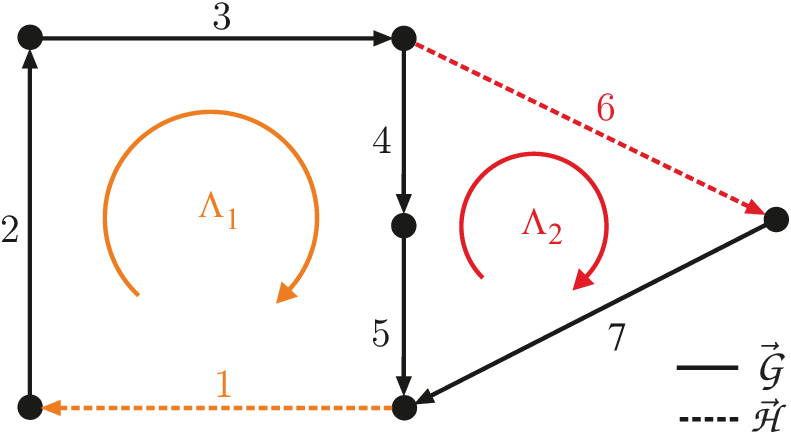}\vspace{%
-3ex}
\end{center}
\caption{a) A 6-bar linkage in its reference configuration. b) Directed
topological graph $\vec{\Gamma}=\vec{\mathcal{G}}\cup \vec{\mathcal{H}}$
with selected $\protect\gamma =2$ FCs $\Lambda _{1},\Lambda _{2}$.}
\label{fig2Loop}
\end{figure}
As an example consider the 6-bar linkage in Fig. \ref{fig2Loop}a). This
example was presented in \cite{CustodioDai2019}. For this linkage, $\gamma
=2 $ FCs can be introduced as in Fig. \ref{fig2Loop}b). The linkage
comprises 7 joints (1 revolute, 4 cylindrical, 1 spherical, 1 prismatic).
All joints are modeled by combination of 1-DOF helical joints. That is, each
cylindrical joint is represented by a revolute followed by a prismatic
joint, and the spherical joint is represented by three successive revolute
joints. The constraint mappings for the FCs $\Lambda _{1},\Lambda _{2}$ are%
\begin{align*}
f_{1}\left( \mathbf{q}\right) :=& \exp (q_{1}\mathbf{Y}_{1})\exp (q_{2}%
\mathbf{Y}_{2})\exp (q_{3,1}\mathbf{Y}_{3,1})\cdot \\
& \exp (q_{3,2}\mathbf{Y}_{3,2})\exp (q_{3,3}\mathbf{Y}_{3,3})\exp (q_{4,1}%
\mathbf{Y}_{4,1})\cdot \\
& \exp (q_{4,2}\mathbf{Y}_{4,2})\exp (q_{5,1}\mathbf{Y}_{5,1})\exp (q_{5,2}%
\mathbf{Y}_{5,2}) \\
f_{2}\left( \mathbf{q}\right) =& \exp (q_{6,1}\mathbf{Y}_{6,1})\exp (q_{6,2}%
\mathbf{Y}_{6,2})\exp (q_{7,1}\mathbf{Y}_{7,1}) \\
& \exp (q_{7,2}\mathbf{Y}_{7,2})\exp (-q_{5,2}\mathbf{Y}_{5,2})\exp (-q_{5,1}%
\mathbf{Y}_{5,1}) \\
& \exp (-q_{4,2}\mathbf{Y}_{4,2})\exp (-q_{4,1}\mathbf{Y}_{4,1})
\end{align*}%
where, for instance, $\mathbf{Y}_{3,1},\mathbf{Y}_{3,2},\mathbf{Y}_{3,3}$
are screw coordinates for the three revolute joints representing the
spherical joint, and $\mathbf{Y}_{4,1},\mathbf{Y}_{4,2}$ those for the
revolute and prismatic joint, respectively, representing the cylindrical
joint.

The \emph{configuration space (c-space)} of the multi-loop linkage is%
\begin{equation}
V:=\left\{ \mathbf{q}\in {\mathbb{V}}^{n}|f_{l}\left( \mathbf{q}\right) =%
\mathbf{I},l=1,\ldots ,\gamma \right\} .  \label{V}
\end{equation}%
defined by the geometric loop constraints $f_{l}\left( \mathbf{q}\right) =%
\mathbf{I}$. A \emph{c-space singularity} is a configuration $\mathbf{q}\in
V $ where $V$ is not a smooth manifold on any neighborhood of $\mathbf{q}$.

The velocity constraints for FC $\Lambda _{l}$ are 
\begin{align}
\sigma _{l,\underline{l}}\dot{q}_{\underline{l}}{\mathbf{S}_{\underline{l}}}%
+\sigma _{l,i}\dot{q}_{i}\mathbf{S}_{i}+\sigma _{l,j}\dot{q}_{j}\mathbf{S}%
_{j}+\ldots +\sigma _{l,{\overline{l}}}\dot{q}_{\overline{l}}\mathbf{S}_{{%
\overline{l}}}=\mathbf{0},&  \label{VelConstr} \\
\ {\underline{l}}\,{<_{l}}\,i\,{<_{l}}\,j\,{<_{l}}\,\ldots {<_{l}}\,%
\overline{l}\in \Lambda _{l}&  \notag
\end{align}%
with the instantaneous screw coordinate vector of joint $i$ 
\begin{equation}
{\mathbf{S}_{i}}%
\hspace{-0.5ex}%
\left( \mathbf{q}\right) :=\mathbf{Ad}_{g_{l,i}\left( \mathbf{q}\right) }%
\mathbf{Y}_{i}  \label{Sil}
\end{equation}%
where%
\begin{equation}
g_{l,i}%
\hspace{-0.5ex}%
\left( \mathbf{q}\right) :=\exp (\sigma _{l,\underline{l}}q_{\underline{l}}%
\mathbf{Y}_{\underline{l}})\cdot \ldots \cdot \exp (\sigma _{l,i-1}q_{i-1}%
\mathbf{Y}_{i-1})\exp (\sigma _{l,i}q_{i}\mathbf{Y}_{i}).  \label{gi}
\end{equation}%
Higher-order loop constraints are given by the time derivatives of the
velocity constraints. For later use, the velocity constraints (\ref%
{VelConstr}) are written as%
\begin{equation}
H_{l}^{\left( 1\right) }%
\hspace{-0.6ex}%
\left( \mathbf{q},\dot{\mathbf{q}}\right) :=\sigma _{l,\underline{l}}\dot{q}%
_{\underline{l}}{\mathbf{S}_{\underline{l}}}%
\hspace{-0.5ex}%
\left( \mathbf{q}\right) +\ldots +\sigma _{l,{\overline{l}}}\dot{q}_{%
\overline{l}}\mathbf{S}_{{\overline{l}}}%
\hspace{-0.5ex}%
\left( \mathbf{q}\right) .  \label{H1}
\end{equation}%
The higher-order constraints are then%
\begin{equation}
H_{l}^{\left( i\right) }%
\hspace{-0.5ex}%
(\mathbf{q},\dot{\mathbf{q}},\ldots ,\mathbf{q}^{\left( i\right) })=\mathbf{0%
}.  \label{ConstrHi}
\end{equation}%
with the mappings $H_{l}^{\left( i\right) }:{\mathbb{V}}^{n}\times {\mathbb{R%
}}^{n}\times \ldots \times {\mathbb{R}}^{n}\rightarrow {\mathbb{R}}^{6}\cong
se\left( 3\right) $ defined as%
\begin{align}
H_{l}^{\left( 2\right) }%
\hspace{-0.6ex}%
\left( \mathbf{q},\dot{\mathbf{q}},\ddot{\mathbf{q}}\right) & :=\frac{d}{dt}%
H_{l}^{\left( 1\right) }%
\hspace{-0.6ex}%
\left( \mathbf{q},\dot{\mathbf{q}}\right)  \notag \\
& \cdots  \label{H} \\
H_{l}^{\left( i\right) }%
\hspace{-0.5ex}%
(\mathbf{q},\dot{\mathbf{q}},\ldots ,\mathbf{q}^{\left( i\right) })& :=\frac{%
d^{i-1}}{dt^{i-1}}H_{l}^{\left( 1\right) }%
\hspace{-0.6ex}%
\left( \mathbf{q},\dot{\mathbf{q}}\right)  \notag
\end{align}%
where $\mathbf{q}^{\left( k\right) }:=\frac{d^{k}}{dt^{k}}\mathbf{q}$. Each $%
H_{l}^{\left( i\right) }$ is a polynomial system of degree $i$ in the
variables $\dot{\mathbf{q}},\ldots ,\mathbf{q}^{\left( i\right) }$.

\section{Local Mobility Analysis%
\label{secLocAnal}%
}

\subsection{Kinematic Tangent Cone%
\label{secKinTangCone}%
}

The \emph{kinematic tangent cone} of $V$ at $\mathbf{q}\in V$ is defined as
the set of tangent vectors to smooth arcs in $V$ passing through $\mathbf{q}$
(denoted $\mathcal{C}_{\mathbf{q}}$):%
\begin{equation}
C_{\mathbf{q}}^{\mathrm{K}}V:=\{\dot{\gamma}|\gamma \in \mathcal{C}_{\mathbf{%
q}}\}\subset {\mathbb{R}}^{n}.  \label{CKV}
\end{equation}%
In other words, it consists of tangents to smooth finite motions. It thus
reveals possible finite motions and the corresponding finite mobility at an
arbitrary (regular or singular) configuration $\mathbf{q}$. At a bifurcation
singularity, $C_{\mathbf{q}}^{\mathrm{K}}V$ is the union of tangent spaces
to the motion branches intersecting at $\mathbf{q}$. Further, at transversal
intersections, it allows to distinguish the motion branches by means of
their tangents. However, at a bifurcation where motion branches of the same
dimension intersect non-transversally, the first-order tangent aspect is not
sufficient to identify the bifurcation.

The concept of tangent cone was first applied to singularity analysis of
linkages in \cite{Lerbet1999}, where $C_{\mathbf{q}}^{\mathrm{K}}V$ was
simply referred to as the tangent cone to $V$ at $\mathbf{q}$. However, as
discussed in the seminal paper by Whitney \cite{Whitney1965}, the cone
defined in (\ref{CKV}) does not capture the local geometry of $V$ if no
smooth arc exists passing through $\mathbf{q}$. In kinematic terminology,
this happens at a deadpoint of a linkage, e.g. at cusp singularity of $V$.
Aiming to describe the local geometry of varieties, there are three relevant
concepts of a tangent cone, the 'tangent semicone', the 'geometric tangent
cone', and the 'algebraic tangent cone' (using an algebraic description of
the linkage kinematics). Only for the algebraic tangent cone there are
method for its actual determination using methods from computational
algebraic geometry \cite{cox-li-os92:IVA}. The latter assume a complex
variety and may not be directly applicable to linkage analysis, however.
Moreover, none of the three versions of tangent cone accounts for general
singularities of the real variety $V$. The cone defined in (\ref{CKV}) does
always reveal the local geometry of finite curves through any $\mathbf{q}$,
i.e. of finite motions through that configuration. Therefore, it is referred
to as the kinematic tangent cone. A detailed discussion can be found in \cite%
{JMR2018,CISMMueller2019,PabloMMT2019}, which also discusses the linkage
reported in \cite{ConnellyServatius1994} exhibiting a cusp singularity.

\subsection{Computational Method}

The kinematic tangent cone is determined by the higher-order loop
constraints \cite{Lerbet1999,JMR2018,CISMMueller2019}. The mappings in (\ref%
{H}) can be determined in closed form by means of Lie brackets (screw
products) of the instantaneous screw coordinates ${\mathbf{S}_{i}}$.
Second-order (acceleration) constraints were reported in \cite{RicoDuffy1996}%
\cite{MMT2014}, third-order (jerk) constraints were reported in \cite%
{Gallardo2001,Gallardo2008,Rico1999,MMT2014}, and fourth-order (jounce)
constraints in \cite{CustodioDai2017}. Such closed form expressions are very
complex and would have to be derived for each order of constraint. Instead
they can be recursively evaluated (symbolically or numerically) in terms of
the instantaneous screw coordinates \cite{MMT2016,Mueller-MMT2019}, which
allows for treatment of general linkages.

Denote the set of vectors satisfying the loop constraints up to order $i$
with%
\begin{equation}
\begin{array}{ll}
K_{\mathbf{q}}^{i}:=\{\mathbf{x}|\exists \mathbf{y},\mathbf{z},\ldots \in {%
\mathbb{R}}^{n}: & H_{l}^{\left( 1\right) }%
\hspace{-0.6ex}%
\left( \mathbf{q},\mathbf{x}\right) =\mathbf{0}, \\ 
& H_{l}^{\left( 2\right) }%
\hspace{-0.6ex}%
\left( \mathbf{q},\mathbf{x},\mathbf{y}\right) =\mathbf{0}, \\ 
& H_{l}^{\left( 3\right) }%
\hspace{-0.6ex}%
\left( \mathbf{q},\mathbf{x},\mathbf{y},\mathbf{z}\right) =\mathbf{0}, \\ 
& \multicolumn{1}{c}{\cdots} \\ 
& \multicolumn{1}{c}{H_{l}^{\left( i\right) }%
\hspace{-0.6ex}%
\left( \mathbf{q},\mathbf{x},\mathbf{y},\mathbf{z,\ldots }\right) =\mathbf{0}%
,l=1,\ldots ,\gamma \}.}%
\end{array}
\label{Ki}
\end{equation}%
The kinematic tangent cone is then determined by the sequence%
\begin{equation}
{C_{\mathbf{q}}^{\text{K}}V}=K_{\mathbf{q}}^{\kappa }\subseteq \ldots
\subseteq K_{\mathbf{q}}^{3}\subseteq K_{\mathbf{q}}^{2}\subseteq {K_{%
\mathbf{q}}^{1}}  \label{CqV}
\end{equation}%
which terminates with a finite $\kappa $. The necessary order depends on the
configuration: $\kappa =\kappa \left( \mathbf{q}\right) $. It is important
to notice that, if $\mathbf{q}$ is a c-space singularity, $K_{\mathbf{q}%
}^{i} $ is not necessarily a vector space but a cone of order $i$, and $C_{%
\mathbf{q}}^{\mathrm{K}}V$ may thus not be a vector space either.

The $K_{\mathbf{q}}^{i}$ can be determined by algebraically eliminating the
variables $\mathbf{y},\mathbf{z,\ldots }$ from the polynomial systems $%
H_{l}^{\left( i\right) }%
\hspace{-0.6ex}%
\left( \mathbf{q},\mathbf{x},\mathbf{y},\mathbf{z,\ldots }\right) =\mathbf{0}
$. When using algorithms from computational algebraic geometry, it must be
ensured that this accounts for the fact that all variables are real.

At a bifurcation singularity $\mathbf{q}$, the kinematic tangent cone splits
into a union of $s\left( \mathbf{q}\right) $ solution sets, denoted $K_{%
\mathbf{q}}^{\kappa \left( \alpha \right) },\alpha =1,\ldots ,s$ (dependence
of $s$ on $\mathbf{q}$ is omitted in the following), 
\begin{equation}
{C_{\mathbf{q}}^{\text{K}}V}=K_{\mathbf{q}}^{\kappa \left( 1\right) }\cup
\ldots \cup K_{\mathbf{q}}^{\kappa \left( s\right) }.  \label{CK}
\end{equation}%
For a transversal intersection, each $K_{\mathbf{q}}^{\kappa \left( \alpha
\right) },\alpha =1,\ldots ,s$ is the tangent space to a motion branch. At
non-transversal intersections, however, there are more than $s$ branches
intersecting at $\mathbf{q}\in V$.

\section{Identification of Non-Transversal Bifurcations via Local Analysis%
\label{secLocAnalTangent}%
}

Tangency of two equidimensional manifolds is characterized by the fact their
tangent spaces are identical. Two equidimensional manifolds are said to be
in higher-order contact, or to possess higher-order tangency, at a common
point $\mathbf{q}$ if their higher-order tangent spaces are identical. That
is, for any smooth curve in one manifold there is a smooth curve in the
other manifold so that the first- and higher-order derivatives of both
curves at $\mathbf{q}$ are parallel. The order up to which this holds, is
referred to as the order of tangency or order of contact. A contact of order
zero is a common point at which there is no common tangent (transveral
intersection). In case of two curves, i.e. 1-dim manifolds, contact of order
one is called 'tangency contact', contact of order two is called 'osculating
contact', and higher order contacts are occasionally called
'superosculation'. Two manifolds of different dimensions, are in tangential
contact if the tangent space of one is a subspace of the tangent space to
other manifold. A formal definition is omitted here and can be found in \cite%
{GuilleminPollack1974,HirschSmale1974}.

\begin{figure}[tb]
\begin{center}
a)~~~~~\hspace{-2ex}\includegraphics[width=9cm]{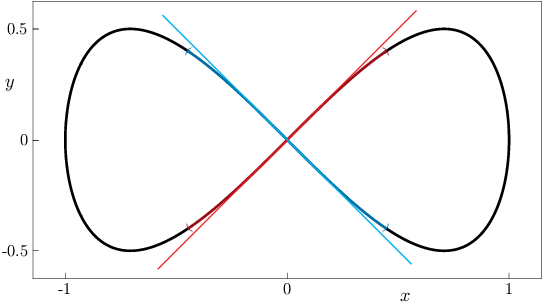}\\[%
0pt]
b)~~~\vspace{-3ex}\includegraphics[width=8.0cm]{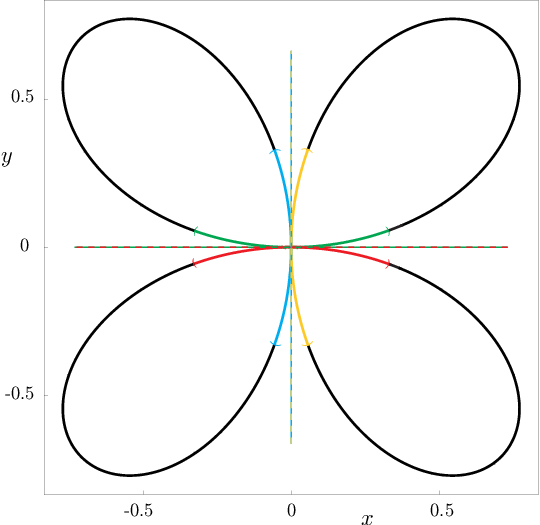}
\end{center}
\caption{a) Eight-curve defined algebraically by $y^{2}=x^{2}-x^{4}$. The
two smooth motion branches (shown in color) intersect tangentially at the
origin. Both belong to the same motion mode (all of the eight-curve). b) The
algebraic curve defined by $\left( x^{2}+y^{2}\right) ^{3}=4x^{2}y^{2}$ is
an example for a c-space exhibiting a bifurcation at the origin with two
tangents of multiplicity 2, leading to tangential as well as transversal
bifurcations of 4 smooth motion branches. The curve represents a single
smooth motion mode.}
\label{figBranches}
\end{figure}

When investigating bifurcations of the c-space, the intersecting manifolds
in question are locally defined in a neighborhood $U\left( \mathbf{q}\right) 
$ of the bifurcation point $\mathbf{q}\in V$ by the equivalence class of
smooth arcs through $\mathbf{q}\in V$. These manifolds are termed 'smooth
motion branches' at $\mathbf{q}\in V$. Notice that the smooth motion
branches may be subvarieties of the same connected component of $V$, as for
the 'eight-curve' and the 'four-leaved clover' in Fig. \ref{figBranches}.
There may be (further) non-smooth motion branches if $\mathbf{q}$ is not
(only) a bifurcation singularity of the c-space $V$.

It should be remarked that a c-space singularity does not necessarily lead
to motion branches. For instance, the planar 5-bar linkage exhibits a
c-space singularity where $V$ consists only of one motion branch \cite%
{JMR2016}. In this case the kinematic tangent cone (and the tangent cone),
and thus locally $V$, is an irreducible cone. 

The motion branches (as smooth manifolds) are distinguished by their
tangential as well as by their higher-order properties. The latter are
already encoded in the solutions $\mathbf{y},\mathbf{z},\ldots $ in (\ref{Ki}%
), which are implicitly determined by the computational steps for
determination of the kinematic tangent cone. In order to utilize this
information, the solution set $\bar{K}_{\mathbf{q}}^{i}\subset {\mathbb{R}}%
^{i\times n}$ of the $i$th-order constraints is introduced as follows%
\begin{equation}
\begin{array}{lll}
\bar{K}_{\mathbf{q}}^{i}:=\{\left( \mathbf{x}_{1},\mathbf{x}_{2},\ldots ,{%
\mathbf{x}}_{i}\right) \in {\mathbb{R}}^{i\times n}: & H^{\left( 1\right) }%
\hspace{-0.6ex}%
\left( \mathbf{q},\mathbf{x}_{1}\right) =\mathbf{0}, & \  \\ 
& H^{\left( 2\right) }%
\hspace{-0.6ex}%
\left( \mathbf{q},\mathbf{x}_{1},\mathbf{x}_{2}\right) =\mathbf{0}, &  \\ 
\multicolumn{1}{c}{} & \multicolumn{1}{c}{\cdots} &  \\ 
\multicolumn{1}{c}{} & \multicolumn{1}{c}{H^{\left( i\right) }%
\hspace{-0.6ex}%
\left( {\mathbf{q}},\mathbf{x}_{1},\mathbf{x}_{2},\ldots ,\mathbf{x}%
_{i}\right) ={\mathbf{0}}\}.} & \ \ \ \ 
\vspace{-8ex}%
\end{array}
\label{Kibar}
\end{equation}%
\vspace{3ex}%

For $1\leq k\leq i$, the projection $\pi _{k}:{\mathbb{R}}^{i\times
n}\rightarrow {\mathbb{R}}^{n}$ to the $k$th factor of $\bar{K}_{\mathbf{q}%
}^{i}$ is%
\begin{equation}
\pi _{k}(\bar{K}_{\mathbf{q}}^{i}):=\{{\mathbf{x}}_{k}\in {\mathbb{R}}%
^{n}|\left( \mathbf{x}_{1},\ldots ,\mathbf{x}_{k},\ldots ,\mathbf{x}%
_{i}\right) \in \bar{K}_{\mathbf{q}}^{i}\}.
\end{equation}%
In particular, the $i$th-order cone (\ref{Ki}) is%
\begin{equation}
K_{\mathbf{q}}^{i}=\pi _{1}(\bar{K}_{\mathbf{q}}^{i}).
\end{equation}%
At a bifurcation of $r$ motion branches, the solution set $\bar{K}_{\mathbf{q%
}}^{i}$ is the union of $s_{i}\leq r$ algebraic varieties, denoted $\bar{K}_{%
\mathbf{q}}^{i\left( \alpha \right) },\alpha =1,\ldots ,s_{i}$,%
\begin{equation}
\bar{K}_{\mathbf{q}}^{i}=\bigcup\limits_{\alpha =1}^{s_{i}}\bar{K}_{\mathbf{q%
}}^{i\left( \alpha \right) }.  \label{Kki}
\end{equation}%
The kinematic tangent cone is the union of $s=s_{\kappa }\leq r$ varieties
in (\ref{CK}). Only for transversal intersections, it is $s=r$. For
non-transversal intersections, however, the tangent spaces to some smooth
motion branches may be identical, so that $s<r$. Clearly $s_{1}=1$ since the
linear system $H_{l}^{\left( 1\right) }\left( \mathbf{q},\mathbf{x}\right) =%
\mathbf{0}$ defines a unique vector space.

Two smooth motion branches, indexed with $\alpha $ and $\beta $, are
distinguished by the fact that there are two solution sets $\bar{K}_{\mathbf{%
q}}^{j\left( \alpha \right) }$ and $\bar{K}_{\mathbf{q}}^{j\left( \beta
\right) }$ in (\ref{Kki}) such that $\pi _{k}(\bar{K}_{\mathbf{q}}^{i\left(
\alpha \right) })\neq \pi _{k}(\bar{K}_{\mathbf{q}}^{i\left( \beta \right)
}) $ for some $i$.

\begin{proposition}
A c-space singularity $\mathbf{q}\in V$ is a \emph{non-transversal
bifurcation point} iff, for some $\alpha ,\beta \in \{1,\ldots ,s_{i}\}$,
there is $k\leq i$ such that%
\begin{eqnarray*}
\pi _{j}(\bar{K}_{\mathbf{q}}^{i\left( \alpha \right) }) &\subseteq &\pi
_{j}(\bar{K}_{\mathbf{q}}^{i\left( \beta \right) }),\forall j=2,\ldots ,k-1
\\
\pi _{j}(\bar{K}_{\mathbf{q}}^{i\left( \alpha \right) }) &\nsubseteq &\pi
_{j}(\bar{K}_{\mathbf{q}}^{i\left( \beta \right) }),\forall j=k,\ldots ,i.
\end{eqnarray*}
\end{proposition}

With slight abuse of notation, the integer $k-1$ is called the \emph{order
of contact} of the motion branches at $\mathbf{q}\in V$. Strict inclusion $%
\pi _{j}(\bar{K}_{\mathbf{q}}^{i\left( \alpha \right) })\subset \pi _{j}(%
\bar{K}_{\mathbf{q}}^{i\left( \beta \right) })$ holds for kinematotropic
mechanisms. From a computational point of view, the analysis boils down to
solving the polynomial systems in the definition (\ref{Kibar}) in terms of
all variable $\mathbf{x}_{1},\ldots ,\mathbf{x}_{\kappa }$. In the rest of
this paper, the method is demonstrated for a few selected examples, where
motion branches of different dimensions intersect.

\section{Examples}

\subsection{2-Loop 6-Bar Linkage}

\begin{figure}[b]
\begin{center}
a)~~~~~\hspace{-2ex}%
\includegraphics[width=6.8cm]{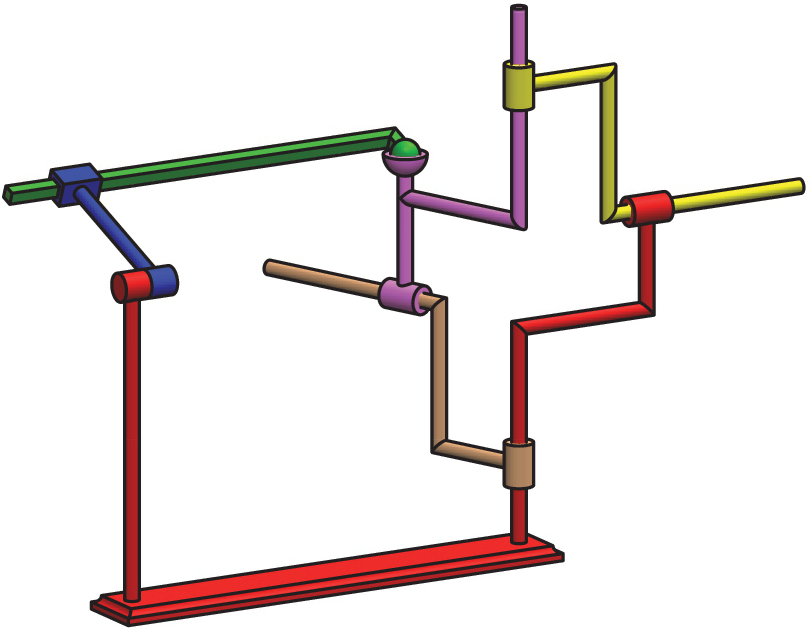}\\[0pt]
b)~~~\includegraphics[width=7cm]{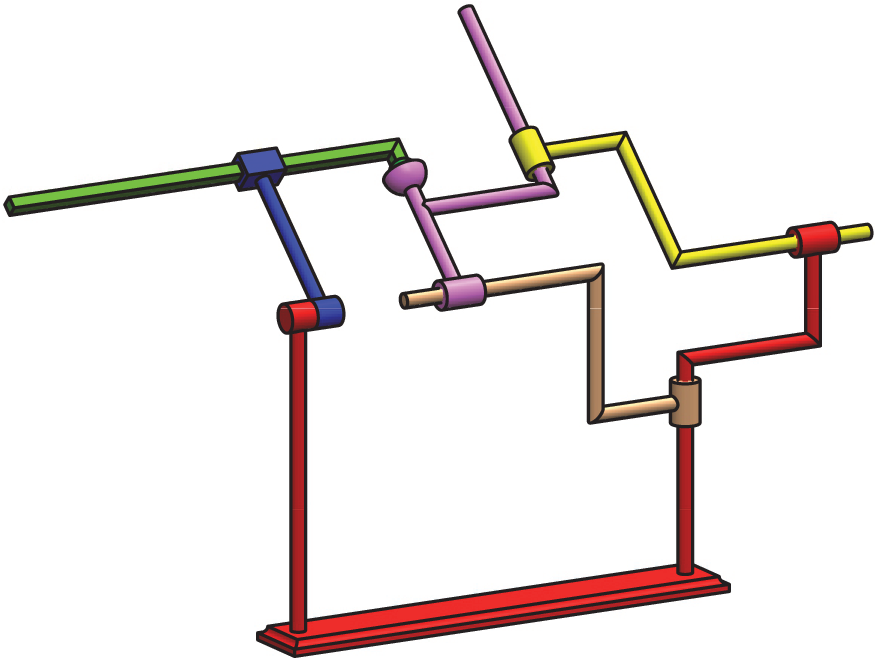}\vspace{%
-3ex}
\end{center}
\caption{The 6-bar linkage in a) configuration with 1 DOF, and b) with 2 DOF}
\label{fig2LoopModes}
\end{figure}
The 6-bar linkage in Fig. \ref{fig2Loop} is analyzed in the shown singular
reference configuration $\mathbf{q}_{0}=\mathbf{0}$. The model comprises
seven joints (1 revolute, 4 cylindrical, 1 spherical, 1 prismatic)
represented by 13 joint screws. Deduced from the shown reference frame, and
with a specific geometry, the screw coordinates are%
\begin{align*}
\mathbf{Y}_{1}& =\left( 1,0,0,0,0,0\right) ^{T},\mathbf{Y}_{2}=\left(
0,0,0,1,0,0\right) ^{T} \\
\mathbf{Y}_{3,1}& =\left( 1,0,0,0,0,-1\right) ^{T},\mathbf{Y}_{3,2}=\left(
0,1,0,0,0,-1\right) ^{T} \\
\mathbf{Y}_{3,3}& =\left( 0,0,1,1,1,0\right) ^{T} \\
\mathbf{Y}_{4,1}& =\left( 1,0,0,0,0,0\right) ^{T},\mathbf{Y}_{4,2}=\left(
0,0,0,1,0,0\right) ^{T} \\
\mathbf{Y}_{5,1}& =\left( 0,1,0,0,0,0\right) ^{T},\mathbf{Y}_{5,2}=\left(
0,0,0,0,1,0\right) ^{T} \\
\mathbf{Y}_{6,1}& =\left( 0,1,0,0,0,0\right) ^{T},\mathbf{Y}_{6,2}=\left(
0,0,0,0,1,0\right) ^{T} \\
\mathbf{Y}_{7,1}& =\left( 1,0,0,0,0,0\right) ^{T},\mathbf{Y}_{7,2}=\left(
0,0,0,1,0,0\right) ^{T}.
\end{align*}%
The first-order cone, determined by the first-order constraints $%
H_{1}^{\left( 1\right) }%
\hspace{-0.6ex}%
\left( \mathbf{q}_{0},\mathbf{x}\right) =H_{2}^{\left( 1\right) }%
\hspace{-0.6ex}%
\left( \mathbf{q}_{0},\mathbf{x}\right) =\mathbf{0}$ with 
\begin{equation*}
\begin{array}{rl}
H_{1}^{\left( 1\right) }\left( {\mathbf{q}},{\mathbf{x}}\right) {=}%
\hspace{-1ex}
& {\mathbf{S}}_{1}x_{1}+\mathbf{S}_{2}x_{2}+\mathbf{S}_{3,1}x_{3,1}+\mathbf{S%
}_{3,2}x_{3,2}+\mathbf{S}_{3,3}x_{3,3}%
\vspace{1ex}
\\ 
& +\mathbf{S}_{4,1}x_{4,1}+\mathbf{S}_{4,2}x_{4,2}+\mathbf{S}_{5,1}x_{5,1}+%
\mathbf{S}_{5,2}x_{5,2}%
\vspace{1ex}
\\ 
H_{2}^{\left( 1\right) }\left( \mathbf{q},{\mathbf{x}}\right) {=}%
\hspace{-1ex}
& \mathbf{S}_{6,1}x_{6,1}+\mathbf{S}_{6,2}x_{6,2}+\mathbf{S}_{7,1}x_{7,2}%
\vspace{1ex}
\\ 
& -\mathbf{S}_{5,2}x_{5,2}-\mathbf{S}_{5,1}x_{5,1}-\mathbf{S}_{4,2}x_{4,2}-%
\mathbf{S}_{4,1}x_{4,1},%
\vspace{1ex}%
\end{array}%
\end{equation*}%
is the 3-dim vector space%
\begin{equation*}
K_{\mathbf{q}_{0}}^{1}=\{\mathbf{x}%
=(s,t,r,-r,0,-s-r,-t,r,0,r,0,-s-r,-t);r,s,t\in {\mathbb{R}}\}.
\end{equation*}%
The second-order cone splits into a 1-dim and a 2-dim vector space%
\begin{equation*}
K_{\mathbf{q}_{0}}^{2}=K_{\mathbf{q}_{0}}^{2\left( 1\right) }\cup K_{\mathbf{%
q}_{0}}^{2\left( 2\right) }
\end{equation*}%
\begin{equation*}
\begin{array}{ll}
\text{where\ } & K_{\mathbf{q}_{0}}^{2\left( 1\right) }=\{\mathbf{x}%
=(s,0,-s,s,0,0,0,-s,0,-s,0,0,0);s\in {\mathbb{R}}\} \\ 
& K_{\mathbf{q}_{0}}^{2\left( 2\right) }=\{\mathbf{x}%
=(s,t,0,0,0,-s,-t,0,0,0,0,-s,-t);s,t\in {\mathbb{R}}\}%
\end{array}%
\end{equation*}%
The computation shows that this is identical to all higher-order cones, and
the kinematic tangent cone is $C_{\mathbf{q}_{0}}^{\mathrm{K}}V=K_{\mathbf{q}%
_{0}}^{2}$.

In order to check whether there are multiple 1-dim or 2-dim motion branches
intersecting tangentially at $\mathbf{q}_{0}$ the solution sets $\bar{K}_{%
\mathbf{q}_{0}}^{i}$ are determined. It turns out that for any order $i>2$,
there is a 1-dim solution set $\bar{K}_{\mathbf{q}_{0}}^{i\left( 1\right) }$
with $K_{\mathbf{q}_{0}}^{i\left( 1\right) }=\pi _{1}(\bar{K}_{\mathbf{q}%
_{0}}^{i\left( 1\right) })$, and a 2-dim solution set $\bar{K}_{\mathbf{q}%
_{0}}^{i\left( 2\right) }$ with $K_{\mathbf{q}_{0}}^{i\left( 2\right) }=\pi
_{1}(\bar{K}_{\mathbf{q}_{0}}^{i\left( 2\right) })$, and $\pi _{j}(\bar{K}_{%
\mathbf{q}_{0}}^{i\left( \alpha \right) })=\pi _{j}(\bar{K}_{\mathbf{q}%
_{0}}^{j\left( \alpha \right) })$ for $j<i,\alpha =1,2$. In other words, the
solutions for any order up to $j$ satisfying the constraints of order $j$
also satisfy the constraints of order $i>j$. By proposition 1, this shows
that $\mathbf{q}_{0}$ is not a non-transversal bifurcation point, and that
there are two motion branches allowing for smooth motions through the
singularity $\mathbf{q}_{0}$. The configuration is a bifurcation point where
the two motion branches intersect transversally, i.e. they are in contact of
order 0. $K_{\mathbf{q}_{0}}^{2\left( 1\right) }$ is the tangent space to a
motion branch in the motion mode where the linkage possesses 1 DOF, and $K_{%
\mathbf{q}_{0}}^{2\left( 2\right) }$ is the tangent space to a motion branch
in the 2 DOF motion mode. Figure \ref{fig2LoopModes} shows a representative
configuration in each mode.

\subsection{Single Loop 7R Linkage}

\begin{figure}[t]
\begin{center}
\includegraphics[width=9.6cm]{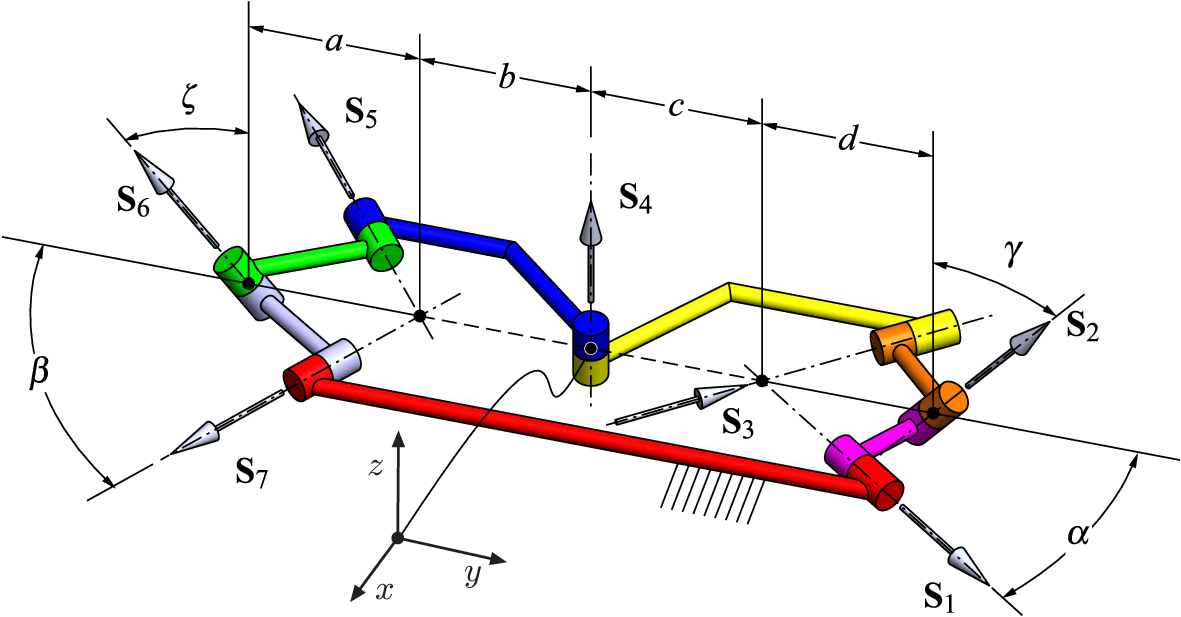}\\[%
0pt]
\end{center}
\caption{7R linkage in the singular configuration $\mathbf{q}_{0}$.}
\label{fig7RTang}
\end{figure}

Figure \ref{fig7RTang} shows a 7R linkage, which is symmetric with respect
to the $yz$-plane according to the shown reference frame. In the shown
reference configuration, the axes of joints 1,3,5, and 7 lie on the $xy$%
-plane, whereas the axes of the remaining joints lie in the $yz$-plane.
Notice that, if joint 4 is removed, the mechanism becomes a 6R
plane-symmetric Bricard mechanism, and thus remains mobile. According to the
design method recently presented in \cite{PabloMMT2020}, inserting joint 4
in the mobile 6R mechanism and restricting its axis to lie in a common plane
with the axes of joint 2 and 6 will lead to a linkage exhibiting
tangentially intersecting motion branches. Observe, however, that joints 2
and 6 are not required to be parallel, as it was in \cite{PabloMMT2020}.

The loop constraints for this 7R single-loop linkage are $f\left( \mathbf{q}%
\right) =\mathbf{I}$, with%
\vspace{-2ex}
\begin{equation*}
f\left( \mathbf{q}\right) =\exp (\mathbf{Y}_{1}q_{1})\exp (\mathbf{Y}%
_{2}q_{2})\cdot \ldots \cdot \exp (\mathbf{Y}_{7}q_{7}).
\end{equation*}%
For the analysis the geometric dimensions are set to $a=b=c=d=2$ and $\alpha
=\gamma =\zeta =\arctan \left( 3/4\right) ,\beta =\arctan \left( 4/3\right) $%
. The screw coordinate vectors are deduced from the reference frame
indicated in Fig. \ref{fig7RTang}%
\begin{align}
\mathbf{Y}_{1}& =\frac{1}{5}\left( 3,4,0,0,0,-6\right) ^{T},\mathbf{Y}_{2}=%
\frac{1}{5}\left( 0,3,4,16,0,0\right) ^{T}  \notag \\
\mathbf{Y}_{3}& =\frac{1}{5}(-3,4,0,0,0,6)^{T},\mathbf{Y}_{5}=-\frac{1}{5}%
(4,3,0,0,0,8)^{T}  \notag \\
\mathbf{Y}_{6}& =\frac{1}{5}(0,-3,4,-16,0,0)^{T},\mathbf{Y}_{7}=\frac{1}{5}%
(4,-3,0,0,0,8)^{T}  \notag \\
\mathbf{Y}_{4}& =(0,0,1,0,0,0)^{T}.
\end{align}%
The first-order cone is the 2-dim vector space%
\begin{equation}
K_{\mathbf{q}_{0}}^{1}=\{\mathbf{x}\in {\mathbb{R}}^{7}|\mathbf{x}%
=(s,t,s,-8/5t,4/3s,t,4/3s),s,t\in {\mathbb{R}}\}.  \label{K1-7R}
\end{equation}%
The second-order cone is the algebraic variety%
\begin{eqnarray}
K_{\mathbf{q}_{0}}^{2} &=&\mathbf{V}%
\big%
(p_{1}\left( {\mathbf{x}}\right) ,p_{2}\left( {\mathbf{x}}\right)
,p_{3}\left( {\mathbf{x}}\right) ,p_{4}\left( {\mathbf{x}}\right)
,p_{5}\left( {\mathbf{x}}\right) ,p_{6}\left( {\mathbf{x}}\right) 
\big%
)  \notag \\
&=&\{\mathbf{x}\in {\mathbb{R}}^{7}|\mathbf{x}=(s,0,s,0,4/3s,0,4/3s),s\in {%
\mathbb{R}}\}
\end{eqnarray}%
determined by the polynomials%
\begin{eqnarray}
p_{1}\left( {\mathbf{x}}\right) &=&4x_{1}-3x_{7},p_{2}\left( {\mathbf{x}}%
\right) =x_{2}-x_{6},p_{3}\left( {\mathbf{x}}\right) =4x_{3}-3x_{7}  \notag
\\
p_{4}\left( {\mathbf{x}}\right) &=&5x_{4}+8x_{6},p_{5}\left( {\mathbf{x}}%
\right) =x_{5}-x_{7},p_{6}\left( {\mathbf{x}}\right) =x_{6}^{2}.
\end{eqnarray}%
It turns out that this is identical to all higher-order cones, i.e. $K_{%
\mathbf{q}_{0}}^{2}=K_{\mathbf{q}_{0}}^{i},i>2$. The kinematic tangent cone
is thus the 1-dim vector space $C_{\mathbf{q}_{0}}^{\mathrm{K}}V=K_{\mathbf{q%
}_{0}}^{2}$. This correctly reveals that there are finite 1-DOF motions
through $\mathbf{q}_{0}$. It also implies that there is either one 1-dim
motion branch through $\mathbf{q}_{0}$ or that there are several
tangentially intersecting motion branches. 

\begin{figure}[t]
\begin{center}
\includegraphics[width=9.1cm]{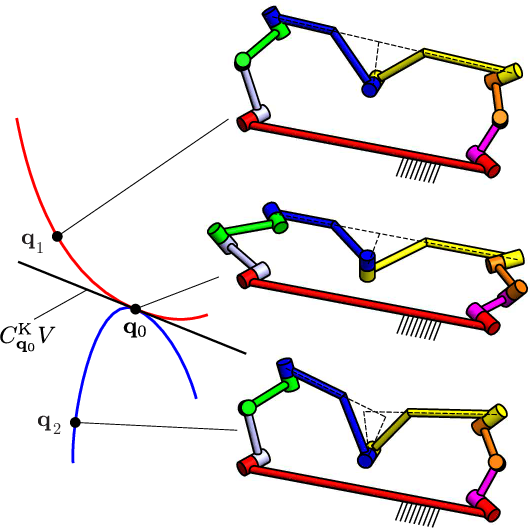}\\[0pt]
\end{center}
\caption{The two motion branches of 7R linkage intersecting tangentially at
the singular configuration $\mathbf{q}_{0}$.}
\label{fig7RModes}
\end{figure}

Up to the 3rd-order there is only one solution branch $\bar{K}_{\mathbf{q}%
_{0}}^{i},i=1,2,3$. For the 4th-order, however, there are two solution
branches, and the solution set splits as (for better readability denoting $%
\mathbf{x}:=\mathbf{x}_{1},\mathbf{y}:=\mathbf{x}_{2},\mathbf{z}:=\mathbf{x}%
_{3},\mathbf{u}:=\mathbf{x}_{4}$)%
\begin{equation}
\bar{K}_{\mathbf{q}_{0}}^{4}=\bar{K}_{\mathbf{q}_{0}}^{4\left( 1\right)
}\cup \bar{K}_{\mathbf{q}_{0}}^{4\left( 2\right) }
\end{equation}%
where one solution set is $\bar{K}_{\mathbf{q}_{0}}^{4\left( 1\right) }=\{(%
\mathbf{x},\mathbf{y},\mathbf{z},\mathbf{u})\}$ with $\mathbf{x}\in C_{%
\mathbf{q}_{0}}^{\mathrm{K}}V$,%
\begin{eqnarray*}
\mathbf{y} &=&%
\Big%
({\small y}_{3}{\small ,-}\frac{43}{30}{\small x}_{1}^{2}{\small ,y}_{3}%
{\small ,0,}\frac{7}{30}{\small x}_{1}^{2}{\small +}\frac{4}{3}{\small y}_{3}%
{\small ,-}\frac{19}{10}{\small x}_{1}^{2}{\small ,}\frac{7}{30}{\small x}%
_{1}^{2}{\small +}\frac{4}{3}{\small y}_{3}%
\Big%
) \\
\mathbf{z} &=&%
\Big%
({\small z}_{1}{\small ,z}_{2}{\small ,z}_{1}{\small ,}\frac{43}{500}{\small %
x}_{1}\left( 43x_{1}^{2}-80{y_{3}}\right) {\small -}\frac{8}{5}{\small z}_{2}%
{\small ,-}\frac{36127}{21600}{\small x}_{1}^{3}{\small +}\frac{7}{10}%
{\small x}_{1}{\small y}_{3} \\
\hspace{-4ex} &&{\small +}\frac{4}{3}{\small z}_{1}{\small ,}\frac{301}{400}%
{\small x}_{1}^{3}{\small -}\frac{7}{5}{\small x}_{1}{\small y}_{3}{\small +z%
}_{2}{\small ,-}\frac{36127}{21600}{\small x}_{1}^{3}{\small +}\frac{7}{10}%
{\small x}_{1}{\small y}_{3}{\small +}\frac{4}{3}{\small z}_{1}%
\Big%
) \\
\hspace{-4ex}\mathbf{u} &=&%
\Big%
({\small u}_{1}{\small ,u}_{2}{\small ,u}_{1}{\small +}\frac{{\small 2}}{%
{\small 375}}{\small x}_{1}\left( {\small 1849x}_{1}^{3}{\small -3440x}_{1}%
{\small y}_{3}{\small -800z}_{2}\right) , \\
&&\frac{{\small 83205x}_{1}^{2}{\small y}_{3}{\small -23264x}_{1}^{4}{\small %
-12900x}_{1}{\small (8z}_{1}{\small +3z}_{2}{\small )-77400y}_{3}^{2}}{%
{\small 11250}}{\small -}\frac{{\small 8}}{{\small 5}}{\small u}_{2}{\small ,%
} \\
\hspace{-4ex} &&\frac{{\small 4}}{{\small 3}}{\small u}_{1}+\frac{{\small %
3523}}{{\small 500}}{\small x}_{1}^{4}-\frac{16049}{720}{\small x}_{1}^{2}%
{\small y}_{3}+\frac{1}{60}x_{1}({\small 56z}_{1}{\small -171z}_{2})+\frac{%
{\small 7}}{{\small 10}}{\small y}_{3}^{2}, \\
\hspace{-4ex} &&{u_{2}}+\frac{7}{9000}\left( 1552x_{1}^{4}+1935x_{1}^{2}{%
y_{3}}-300x_{1}(8{z_{1}}+3{z_{2}})-1800{y_{3}^{2}}\right) , \\
\hspace{-4ex} &&\frac{4{u_{1}}}{3}-\frac{7}{3600}180x_{1}^{4}+4387x_{1}^{2}{%
y_{3}}-60x_{1}(8{z_{1}}+3{z_{2}})-360{y_{3}^{2}}%
\Big%
)
\end{eqnarray*}%
and the other solution set is $\bar{K}_{\mathbf{q}_{0}}^{4\left( 2\right)
}=\{(\mathbf{x},\mathbf{y},\mathbf{z},\mathbf{u})\}$ with $\mathbf{x}\in C_{%
\mathbf{q}_{0}}^{\mathrm{K}}V$,%
\begin{eqnarray*}
\hspace{-2ex}%
\mathbf{y} &=&%
\Big%
({y_{3}},-\frac{3}{5}x_{1}^{2},{y_{3}},-\frac{4}{3}x_{1}^{2},\frac{7}{30}%
x_{1}^{2}+\frac{4}{3}{y_{3}},-\frac{16}{15}x_{1}^{2},\frac{7}{30}x_{1}^{2}+%
\frac{4}{3}{y_{3}}%
\Big%
) \\
\hspace{-2ex}%
\mathbf{z} &=&%
\Big%
({z_{1}},{z_{2}},{z_{1}}-\frac{8}{3}x_{1}^{3},\frac{43}{250}x_{1}\left(
9x_{1}^{2}-40{y_{3}}\right) -\frac{8}{5}{z_{2}}, \\
\hspace{-2ex}
&&-\frac{37301}{10800}x_{1}^{3}+\frac{7}{10}x_{1}{y_{3}}+\frac{4}{3}{z_{1}},%
\frac{63}{200}x_{1}^{3}-\frac{7}{5}x_{1}{y_{3}}+{z_{2}}, \\
\hspace{-2ex}
&&-\frac{15701}{10800}x_{1}^{3}+\frac{7}{10}x_{1}{y_{3}}+\frac{4}{3}{z_{1}}%
\Big%
) \\
\hspace{-2ex}%
\mathbf{u} &=&%
\Big%
({u_{1}},{u_{2}},{u_{1}}+\frac{2}{375}x_{1}\left( 999x_{1}^{3}-4440x_{1}{%
y_{3}}-800{z_{2}}\right) , \\
\hspace{-2ex}
&&\frac{40543x_{1}^{4}+17415x_{1}^{2}{y_{3}}-6450x_{1}(8{z_{1}}+3{z_{2}}%
)-38700{y_{3}^{2}}-9000{u_{2}}}{5625}, \\
\hspace{-2ex}
&&\frac{4}{3}{u_{1}}+\frac{25639x_{1}^{4}-232675x_{1}^{2}{y_{3}}+150x_{1}(56{%
z_{1}}-171{z_{2}})+6300{y_{3}^{2}}}{9000}, \\
\hspace{-2ex}
&&{u_{2}}+\frac{7}{4500}\left( 1801x_{1}^{4}+405x_{1}^{2}{y_{3}}-150x_{1}(8{%
z_{1}}+3{z_{2}})-900{y_{3}^{2}}\right) , \\
\hspace{-2ex}
&&\frac{4}{3}{u_{1}}-\frac{7}{1800}\left( 295x_{1}^{4}+2081x_{1}^{2}{y_{3}}%
-30x_{1}(8{z_{1}}+3{z_{2}})-180{y_{3}^{2}}\right) 
\Big%
).
\end{eqnarray*}%
Consequently, there are two motion branches that are intersecting
tangentially, and $C_{\mathbf{q}_{0}}^{\mathrm{K}}V$ is the (common) tangent
space to these motion branches, since $\pi _{1}(\bar{K}_{\mathbf{q}%
_{0}}^{4\left( 1\right) })=\pi _{1}(\bar{K}_{\mathbf{q}_{0}}^{4\left(
2\right) })=C_{\mathbf{q}_{0}}^{\mathrm{K}}V$. This is schematically shown
in Fig. \ref{fig7RModes} where a configuration $\mathbf{q}_{1}$ and $\mathbf{%
q}_{2}$, respectively, in each motion branch is shown.

{Note that in both cases $x_{4}=0$. However, $y_{4}=0$ for $\mathbf{y}\in
\pi _{2}(\overline{K}_{\mathbf{q}_{0}}^{4(1)})$, while $y_{4}\neq 0$ for $%
\mathbf{y}\in \pi _{2}(\overline{K}_{\mathbf{q}_{0}}^{4(2)})$. This proves
that there are two different branches of motion intersecting tangentially at 
$\mathbf{q}_{0}$ with the same joint velocities but different joint
accelerations. }Moreover, $\pi _{i}(\bar{K}_{\mathbf{q}_{0}}^{4\left(
1\right) })\neq \pi _{i}(\bar{K}_{\mathbf{q}_{0}}^{4\left( 2\right)
}),i=2,3,4$, and thus these two branches are in first-order contact. This
implies that the linkage can transit between the two motion branches via $%
\mathbf{q}_{0}$ by a first-order continuous motion.

It is instructive to explain the behavior of the linkage in each branch. In
branch 2, although $x_{4}=0$, the fact that $y_{4}\neq 0$ shows that joint 4
is not inactive but in a dead point and $x_{4}\neq 0$ in the neighborhood of 
$\mathbf{q}_{0}$. Hence, in branch 2, the mechanism behaves as a general 7R
linkage with 1 DOF. In branch 1, both $x_{4}$ and $y_{4}$ are equal to 0,
therefore it cannot be concluded whether joint 4 is active or inactive.
However, since for this branch $\mathrm{\dim }K_{\mathbf{q}_{0}}^{4(1)}=1$,
and it is known that $\mathrm{\dim }K_{\mathbf{q}_{0}}^{i(1)}\leq \dim K_{%
\mathbf{q}_{0}}^{j(1)},\forall i>j$, the mobility of the mechanism in this
branch is either 1 or 0. Moreover, at $\mathbf{q}_{0}$ the mechanism is in a
plane-symmetric configuration, and removal of joint 4 at $\mathbf{q}_{0}$
would lead to a 6R plane-symmetric linkage, which is known from Bricard to
be movable with 1 DOF. Therefore, a 1-dimensional branch where joint 4 is
inactive and the linkage remains plane symmetric must intersect $\mathbf{q}%
_{0}$. Since in branch 2, joint 4 is active, the only possibility is that in
branch 1 the linkage resembles a 6R Bricard mechanism. This also indicates
that not only $x_{4}$ and $y_{4}$ but also all higher-order derivatives
vanish at $\mathbf{q}_{0}$.

\section{Conclusion}

A generally applicable approach to the identification of non-transversal
bifurcations was presented. The approach exploits relations already
available from the computational steps for the construction of the kinematic
tangent cone. Thus all necessary computational steps can be performed
efficiently either symbolically or numerically. This result complements the
higher-order local mobility and singularity analysis in that it provides
additional information characterizing the finite mobility at a given
configuration.

Beside the theoretical interest, mechanisms exhibiting non-transversal, and
in particular tangential, bifurcations of motion branches may potentially
have practical applications. A $k$th-order tangential intersection implies
that the mechanism can transit from one motion branch to another by
performing a $C^{k}$ continuous motion, i.e. a motion with continuous
derivatives $\dot{\mathbf{q}},\ldots ,\mathbf{q}^{\left( k\right) }$. This
is certainly advantageous for the control of such mechanisms. A particularly
interesting type of tangential intersections can be expected in case of
kinematotropic mechanisms. The application of the proposed method to such
linkages cannot be presented in this paper as the intermediate steps are too
complex. Kinematotropic mechanisms have not yet seen wider practical
applications \cite{BuchtaVoglewedeDETC2017}, but the ability to change
motion branches with certain continuity may be a further beneficial feature.
The presented approach provides conditions for $k$th-order continuity of the
transitory motions between the motion branches of different dimensions that
can be incorporated in the design of kinematotropic mechanisms \cite%
{QinDaiGago2014,Kong-MMT2018,GallettiGiannotti-DETC2002,Ibarreche-Meccanica2019,LeeHerve2005}

\section*{Acknowledgement}

The first author acknowledges support by the LCM-K2 Center within the
framework of the Austrian COMET-K2 program. The second and third authors
acknowledge the support by the Engineering and Physical Science Research
Council (EPSRC) of the UK with grant numbers EP/P026087/1 and EP/S019790/1.

\bibliographystyle{IEEEtran}
\bibliography{TangIntersect_IDETC2020}

\end{document}